\documentclass[10pt,twocolumn,letterpaper]{article}

\usepackage{cvpr}              
\definecolor{cvprblue}{rgb}{0.21,0.49,0.74}
\usepackage{multirow}
\usepackage{caption}
\usepackage[pagebackref,breaklinks,colorlinks,allcolors=cvprblue]{hyperref}
\usepackage[linesnumbered,ruled,vlined]{algorithm2e}
\usepackage{algorithmic}
\usepackage{booktabs}
\usepackage{tcolorbox} 
\definecolor{task}{RGB}{139, 0, 0}
\definecolor{reward}{RGB}{46,139,87}
\definecolor{explore}{RGB}{218,165,32}
\definecolor{update}{RGB}{0, 0, 0}
\newcommand{\Model}{EEA\xspace}
\def\paperID{7664} 
\def\confName{CVPR}
\def\confYear{2026}
\title{EEA: Exploration–Exploitation Agent for Long Video Understanding}

\author{
Te Yang$^{1,2}$,
Xiangyu Zhu$^{1,2}$,
Bo Wang$^{3}$,
Quan Chen$^{3}$,
Peng Jiang$^{3}$,
Zhen Lei$^{1,2,4}$\\
$^{1}$State Key Laboratory of Multimodal Artificial Intelligence Systems, \\
Institute of Automation, Chinese Academy of Sciences\\
$^{2}$School of Artificial Intelligence, University of Chinese Academy of Sciences, Beijing, China\\
$^{3}$Kuaishou Technology, Beijing, China\\
$^{4}$Centre for Artificial Intelligence and Robotics, HKISI, Chinese Academy of Sciences, Beijing, China\\[4pt]
{\tt\small \{yangte2021, xiangyu.zhu, zhen.lei\}@ia.ac.cn}\\
{\tt\small wangbo0060@163.com, myctllmail@163.com, jp2006@139.com}
}

\begin{document}
\maketitle
\begin{abstract}
Long-form video understanding requires efficient navigation of extensive visual data to pinpoint sparse yet critical information.
Current approaches to long-form video understanding either suffer from severe computational overhead due to dense preprocessing, or fail to effectively balance exploration and exploitation, resulting in incomplete information coverage and inefficiency.
In this work, we introduce \Model, a novel video agent framework that archives exploration-exploitation balance through semantic guidance with hierarchical tree search process.
\Model autonomously discovers and dynamically updates task-relevant semantic queries, and collects video frames closely matched to these queries as semantic anchors. 
During the tree search process, instead of uniform expansion, \Model preferentially explores semantically relevant frames while ensuring sufficient coverage within unknown segments. 
Moreover, \Model adaptively combines intrinsic rewards from vision-language models (VLMs) with semantic priors by explicitly modeling uncertainty to achieve stable and precise evaluation of video segments.
Experiments across various long-video benchmarks validate the superior performance and computational efficiency of our proposed method.          
\end{abstract}    
\section{Introduction}
\label{sec:intro}
Long-form video understanding~\cite{wu2021towards, soldan2022mad} remains a fundamental challenge in computer vision and artificial intelligence, forming the cornerstone of applications that require long-term temporal reasoning over visual contents, such as documentary analysis, sports understanding, and live-stream comprehension.
Long videos are often characterized by highly uneven information density and complex event structures~\cite{tapaswi2016movieqa, wu2024longvideobench}.
Extracting valuable information from such videos requires models to identify relevant information from massive redundancy and handle long-term temporal dependencies~\cite{li2024mvbench,  wang2024qwen2, bai2025qwen2}.

\begin{figure}[t]
    \centering
    \centerline{
    \includegraphics[width=\linewidth]{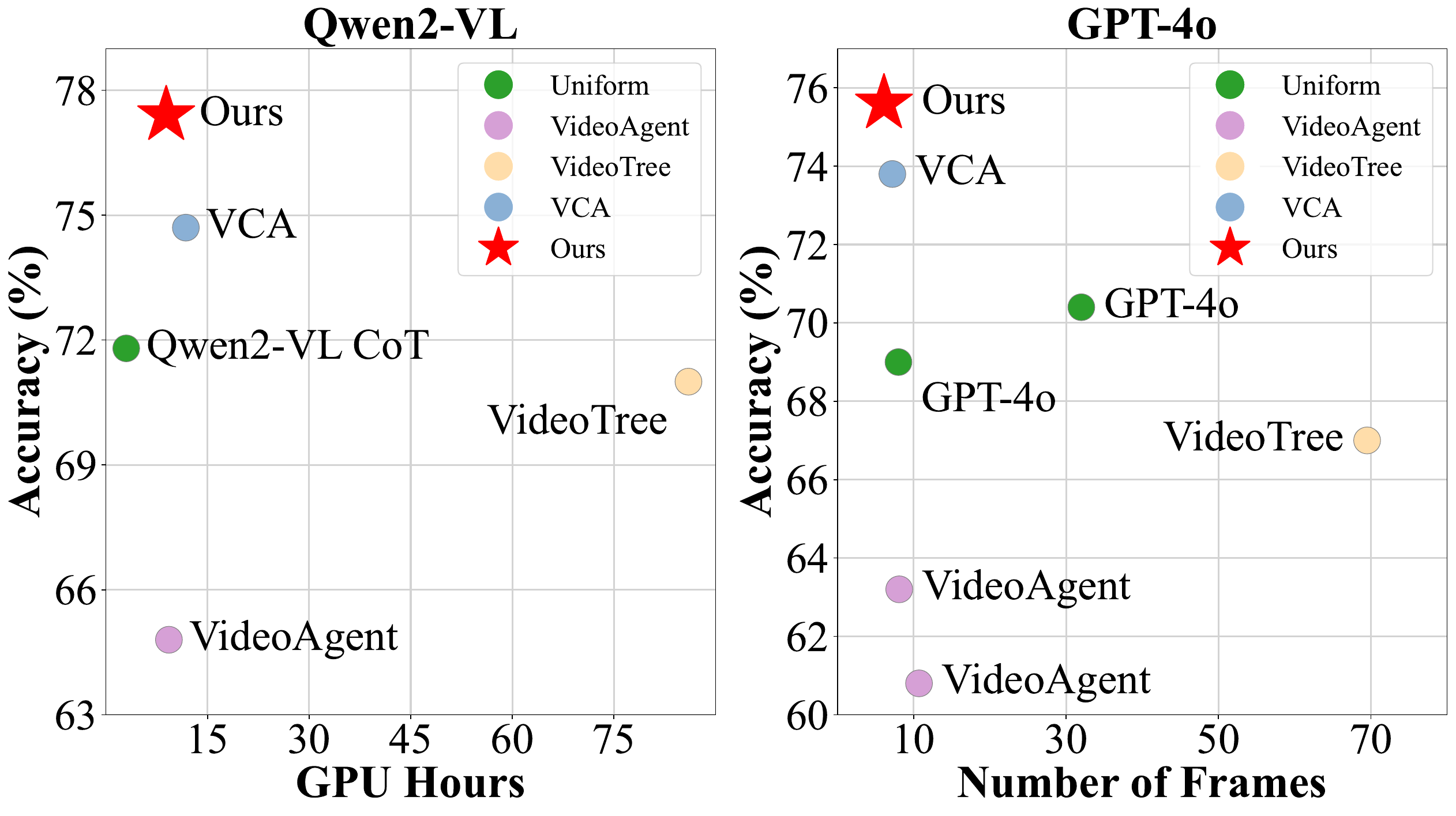}}
    \vspace{-0.5em}
    \caption{\textcolor{update}{\textbf{Accuracy versus. Computational Cost and Frame Utilization.}
Our agent achieves higher accuracy with improved efficiency, requiring fewer observed frames and comparable or lower GPU cost on EgoSchema, under both GPT-4o and Qwen2-VL-72B settings.}}
    \label{fig:teaser}
    \vspace{-1.7em}
\end{figure}

Existing VLM-based methods~\cite{team2024gemini, zhang2024long} rely on dense frame sampling, resulting in redundant and inefficient representations~\cite{wang2024lvbench,yu2024self}.
Meanwhile, agent systems that require costly preprocessing such as dense captioning~\cite{luo2024videorag, ma2024drvideo, zhang2025deep, pang2025mr,fan2024videoagent} often incurs substantial computational overhead.
These challenges call for an agent capable of reasoning selectively and efficiently, acquiring only the most informative frames for decision-making.

Recent long video understanding studies have introduced an on-the-fly agent-based paradigm~\cite{kugo2025videomultiagents, wang2024videotreeadaptivetreebasedvideo}, which formulates long-video understanding as a online decision-making process, avoiding costly offline preprocessing.
Exploitation based methods~\cite{VideoAgent,yuan2025videodeepresearch, ye2025re} use existing queries as a guide for action, continuously refining the search process through iterative retrieval to ultimately aggregate only the most relevant information required.
However, their heavy reliance on exploitation tends to trap the search in local optima, neglecting novel evidence beyond initial queries.
Exploration based methods~\cite{yang2025vca, wang2025videochat}, improve long-video reasoning through hierarchical tree search, where the video is recursively partitioned into multiple segments represented as tree nodes. 
At each search step, a node is uniformly expanded into finer-grained child nodes, and the most promising nodes are selected by comparing their relevance to the user instructions.
The key challenge lies in determining how to expand nodes effectively. 
To ensure efficiency, existing methods typically sample only sparse frames within each node, which serve both as expansion boundaries and as references for deciding subsequent actions.
However, under uniform partitioning, this sparse sampling often results in critical information being overlooked.

\begin{figure}[t]
    \centering
    \includegraphics[width=\linewidth]{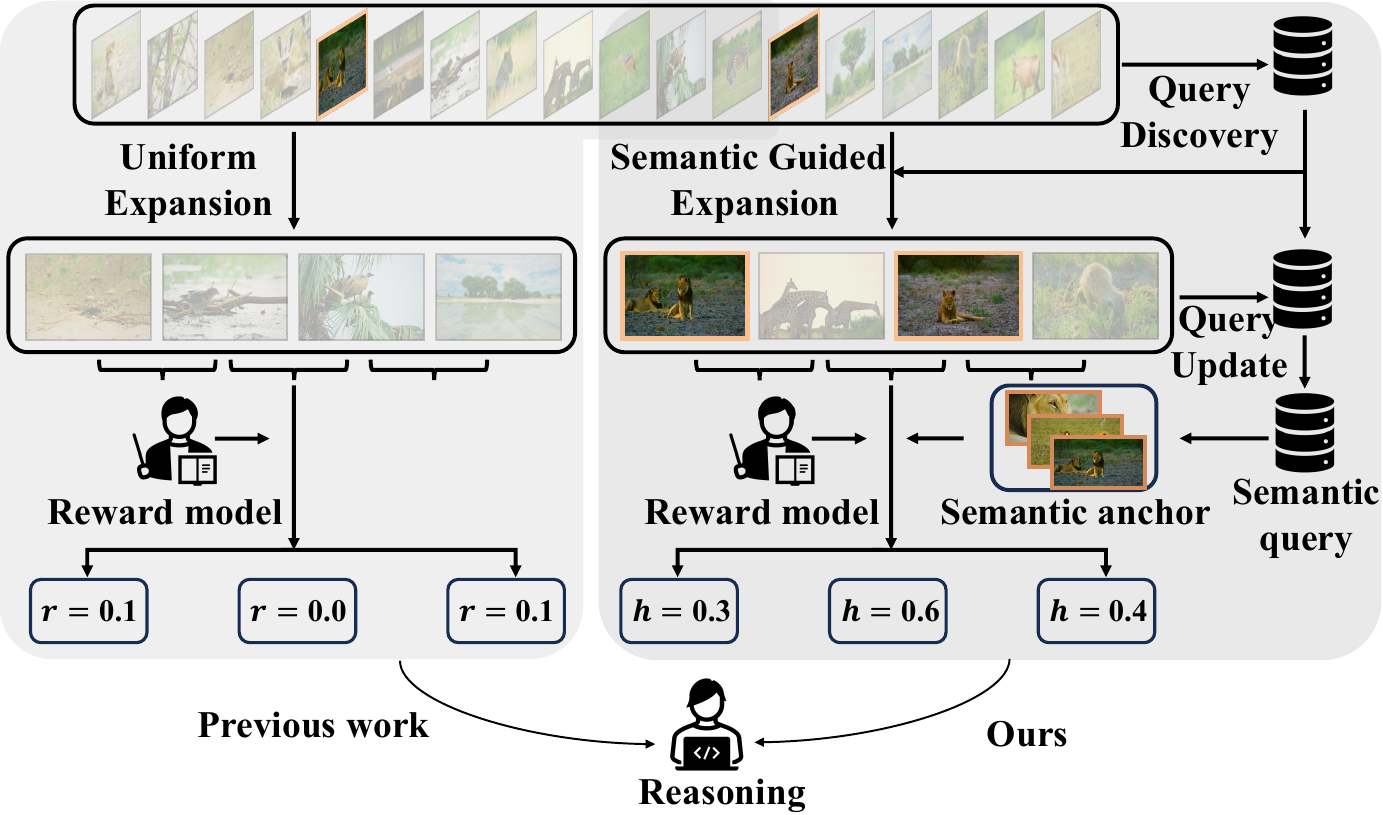}
    \vspace{-1.5em}
    \caption{\textcolor{update}{\textbf{Comparison of \Model with prior works.} 
    \textbf{Difference-1:}
    Prior methods use no semantic priors. \Model performs dynamic query discovery and query update during exploration, enabling progressively refined guidance over long videos.
    \textbf{Difference-2:} \Model performs semantic-guided expansion, leveraging semantic queries and anchors to focus on relevant frames. In contrast, previous methods rely on blind uniform sampling, risking the omission of critical events.
    \textbf{Difference-3:}: Previous methods base exploration decisions solely on a potentially noisy intrinsic reward. \Model enhances more stable evaluation by fusing this reward with a robust query-based score.
    }}
    \label{fig:framework-compare}
    \vspace{-1.8em}
\end{figure}

In this work, we introduce \Model, an \textbf{E}xploration–\textbf{E}xploitation \textbf{A}gent for long-form video understanding motivated by human cognitive processes.
When humans attempt to understand an unfamiliar environment, they typically do not explore blindly. Instead, guided by initial hypotheses based on task semantics, they iteratively balance exploration of new information and exploitation of existing knowledge, continuously refining their understanding~\cite{chakroun2020dopaminergic,wyatt2024exploration}.
Inspired by this processes, we propose a novel video agent framework, \Model, that efficiently leverages existing textual information while actively exploring unknown regions, enabling accurate and efficient reasoning over long-form videos.
\Model~is built upon a hierarchical tree search framework, focusing on three critical steps of the search process: maintaining semantic queries, evaluating nodes for expansion, and strategically expanding nodes into child segments.
(1) To make sure that our semantic queries are not static but dynamically updated, we propose Dynamic Query Management(DQM) to continuously discover and enrich additional semantic priors during exploration. By dynamically generating new semantic queries from the visual information of the selected node in each iteration, \Model effectively addresses the insufficiency of semantic priors derived from initial user instructions.
(2) To determine which node to expand, we introduce a sophisticated reward mechanism named Uncertainty-Aware Reward Fusion (UARF). This reward adaptively integrates intrinsic signals from the VLM with query scores derived from semantic priors, effectively addressing the insufficient discriminability that arises when relying solely on intrinsic rewards.
(3) Instead of uniformly partitioning nodes into child segments, we propose Semantic-Guided Expansion(SGE) by strategically expanding node guided by semantic queries, effectively balancing focused search and broad information coverage. 
The key differences between \Model and the prior work are illustrated in Fig \ref{fig:framework-compare}.
Extensive experiments conducted across multiple benchmarks validate that our proposed \Model framework substantially enhances both the effectiveness and efficiency of agents in long-form video reasoning tasks, achieving superior performance compared to baseline approaches. 
Our contributions can be summarized as fourfold:
\begin{itemize}
    \item We propose \Model, a \textbf{E}xploration–\textbf{E}xploitation agent framework that achieves an effective balance between exploring new information and leveraging known semantics for efficient long-form video reasoning.
    \item \Model employs a novel Semantic Guided Expansion approach in hierarchical tree search process that effectively balances targeted search with broad coverage, significantly improving long-form video understanding.
    \item \Model designs an Uncertainty-Aware Reward Fusion mechanism that integrates intrinsic reward signals with semantic priors, providing  stable and discriminative evaluation during long-horizon reasoning.
    \item \Model introduces a Dynamic Query Mangement mechanism that continuously refines semantic queries throughout reasoning, effectively resolving the insufficiency of  initial queries and enhancing the agent’s ability to discover and accurately identify critical evidence.
\end{itemize}

\begin{figure*}[htbp]
    \centering
    \vspace{-0.8em}
    \includegraphics[width=\linewidth]{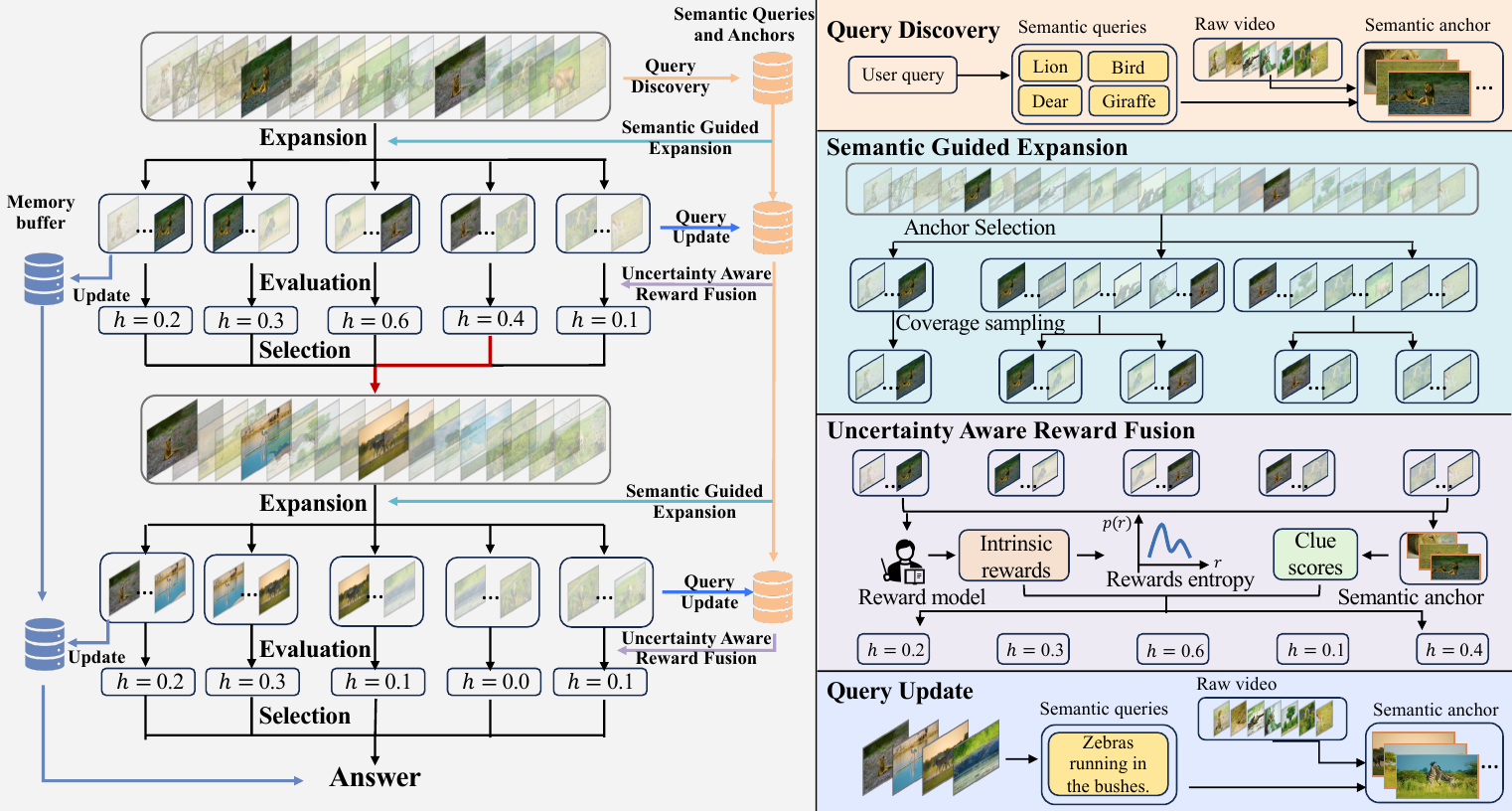}
    \caption{\textbf{Pipeline of \Model Framework.} The agent first derives semantic queries from the query and identifies their corresponding semantic anchors in the video. Guided by these anchors, it performs semantic-guided expansion to expand candidate nodes. It then evaluates each node with a fused reward obtained through uncertainty-aware reward fusion, combining intrinsic reward and query score. Then the agent decides its next action based on obtained information and updates the semantic queries, anchors, and memory buffer.}
    \label{fig:framework}
    \vspace{-0.2em}
\end{figure*}

\section{Related work}
\label{sec:related_works}
\subsection {Large Vision-Language Models}
Large vision-language models (VLMs), empowered by recent advances in large language models (LLMs)~\cite{achiam2023gpt, touvron2023llama, naveed2025comprehensive}, have demonstrated remarkable performance across various multimodal tasks, including visual question answering, image captioning, and multimodal reasoning~\cite{wang2024qwen2, bai2025qwen2, team2024gemini}.
However, applying these VLMs directly to video understanding, especially long-form videos, faces significant computational challenges. The common practice of uniformly sampling frames (e.g., one frame per second) causes the computational burden to grow quadratically as video length increases, severely limiting their efficiency and scalability.
To overcome this limitation, we propose the \Model framework, which leverages semantic-guided hierarchical tree search to effectively manage the context length and selectively attend to critical visual content, significantly reducing computational overhead and improving reasoning accuracy over long videos.

\subsection{Long Video Understanding}
Early video understanding approaches mainly rely on two-stream networks~\cite{feichtenhofer2016convolutionaltwostreamnetworkfusion,Feichtenhofer_2019_ICCV,wang2016temporalsegmentnetworksgood,ng2015shortsnippetsdeepnetworks, kangaspunta2021adaptive, vaidya2022co}  and 3D CNNs~\cite{ji2012-3d-pami, karpathy2014large,tran2015learningspatiotemporalfeatures3d,carreira2018quovadisactionrecognition,xie2018rethinkingspatiotemporalfeaturelearning,brattoli2020rethinking} for spatio-temporal feature extraction. 
The introduction of self-attention mechanisms~\cite{9710415,fan2021multiscalevisiontransformers,bertasius2021spacetimeattentionneedvideo} and self-supervised pretraining~\cite{sun2019videobertjointmodelvideo, zhu2020actbertlearninggloballocalvideotext, tong2022videomaemaskedautoencodersdataefficient} further enhances the model’s capacity for temporal modeling and cross-task generalization.
Recently, large vision-language models (LVLMs) ~\cite{hong2024cogvlm2visuallanguagemodels,Qwen2VL,zhang2024llavanextvideo} have demonstrated strong capabilities in video understanding tasks. However, the long-context nature of long-form video understanding poses severe challenges to model efficiency. 
Sparse representation strategies construct compact visual representations by pruning and merging tokens~\cite{song2024moviechat, jin2024chat, alvar2025divprune} or selecting key frames~\cite{ren2023timechat,li2024llms, xu2023retrievalbased, han2025dynfocus}, often at the cost of significant information loss.
Other methods~\cite{zhang2023simple,romero2024question,luo2024videorag, ma2024drvideo, zhang2025deep, pang2025mr} obtain structured representations of videos by generating textual descriptions of visual content, which, however, incurs substantial computational overhead.

On-the-fly agent-based methods~\cite{kugo2025videomultiagents, wang2024videotreeadaptivetreebasedvideo, VideoAgent,yuan2025videodeepresearch, ye2025re, yang2025vca, wang2025videochat}, instead of preprocessing the whole video in advance, dynamically acquire the information required for reasoning during the agent’s interaction process. VideoAgent~\cite{VideoAgent} retrieves and integrates visual evidence with the help of VLM and CLIP tools, while VideoDeepResearch~\cite{yuan2025videodeepresearch} leverages a text-only reasoning model to iteratively retrieve and analyze relevant video segments through multimodal tool use. VideoTree~\cite{wang2024videotreeadaptivetreebasedvideo} adopts a coarse-to-fine hierarchical exploration strategy that expands nodes using relevance-based hierarchical levels and VCA~\cite{yang2025vca} introduces a curiosity-driven video agent that performs tree-search-based self-exploration guided by intrinsic rewards. However, all these methods focus exclusively on either exploring unseen information or exploiting existing knowledge. 
In this paper, we propose \Model, a novel agent framework for long video understanding. By fully leveraging task semantics to identify key frames while sampling from unexplored segments, \Model achieves a balanced trade-off between exploration and exploitation, enhancing both the performance and efficiency of long video understanding.

\section{Method}
\label{sec:method}
Inspired by human cognitive processes, we introduce \Model, a long-video understanding agent that mimics the iterative cycle of exploration and evidence gathering. \Model navigates the vast search space of long videos via a novel tree-search framework that dynamically balances exploitation of known clues with exploration of unknown regions.

Our approach is built upon three key contributions. First, we propose a \emph{Dynamic Query Management } mechanism that not only extracts initial semantic queries from the user instructions but also continuously refines them based on newly observed visual evidence. Second, to overcome the limited sensitivity of standard reward models, we design an \emph{Uncertainty-Aware Reward Fusion} technique. This method adaptively integrates intrinsic VLM rewards with an additional query score, significantly enhancing the discriminability of segment evaluations. Finally, these components are unified within a \emph{Semantic Guided Tree Search} algorithm. The key innovation of this search is our \emph{Semantic-Guided Expansion} strategy, which strategically prioritizes sampling around high-relevance semantic anchors while ensuring broad temporal coverage. This enables efficient localization of key segments while minimizing redundant exploration.
In this section, we first elaborate on how semantic queries are dynamically managed (Section~\ref{sec:clue_management}). Next, we introduce our uncertainty-aware reward fusion mechanism (Section~\ref{sec:Uncertainty-Aware-Reward-Fusion}). Finally, we detail the semantic guided tree search process and its core expansion strategy (Section~\ref{sec:tree-search}).

\subsection{Dynamic Query Management}
\label{sec:clue_management}
In long-video understanding tasks, agents typically face enormous search spaces and sparse distributions of critical information~\cite{ye2025re,mangalam2023egoschema,qian2024streaming}, often resulting in aimless exploration of numerous irrelevant segments.
Although existing methods leverage semantic queries to reduce search complexity~\cite{zou2024language, zala2023hierarchical}, these approaches typically focus only on static initialization, neglecting the dynamic emergence of new queries during exploration.
Inspired by the human cognitive mechanism of dynamically updating attentional focus during video viewing, we introduce a dynamic query management mechanism.
Specifically, we not only effectively extract initial semantic queries from the user instruction at the beginning of exploration but also incrementally refine and update these queries based on newly observed visual evidence, thereby effectively guiding the agent to focus on more relevant video segments.

Given a user instruction $Q$, we first extract an initial set of semantic queries $\mathcal{C}_0=\{c_i\}$.
These semantic queries typically encompass key elements from the instruction, such as objects, scenes, and actions.
Then, for each semantic query $c_i$, we perform cross-modal retrieval to obtain a set of candidate clips $\mathcal{D}_i=\{(t_{k}^{(i)},\phi_{k}^{(i)})\}$, each with an associated timestamp $t_{k}^{(i)}$ and semantic similarity score $\phi_{k}^{(i)}$.
To reduce redundancy, we cluster these segments into multiple clusters based on their temporal overlaps.
Finally, within each cluster, we select the frame with the highest semantic similarity score to form the semantic anchor set $\mathcal{A}_0=\{(f_{A}^{m}, \phi^{m})\}$.

After the $t$-th round of exploration, the agent observes a set of sampled frames $\mathcal{F}_t$ from the selected video segment.
Based on these newly observed visual cues and the previously collected queries $\mathcal{C}_{t-1}$, we extract additional semantic queries $\Delta \mathcal{C}_t$ exclusively derived from the current observations. 
These new queries are then merged with historical queries to form an updated query set.
Subsequently, we use these newly added queries $\Delta \mathcal{C}_t$ to acquire an incremental set of candidate video segments $\Delta \mathcal{D}_t$ and merge this incremental set $\Delta \mathcal{D}_t$ with the previous candidate clip set $\mathcal{D}_{t-1}$, removing any duplicate segments retrieved by different queries.
Finally, these refined segments are re-clustered based on their temporal overlaps to generate an updated set of semantic anchors $\mathcal{A}_t$.

In the video exploration process, these semantic anchors serve as explicit cues that help the agent locate relevant segments within very long videos.
Intuitively, these anchors function like landmarks on an unknown map, effectively providing exploration directions and preventing aimless searching.
By offering clear semantic references, the dynamically updated anchors significantly enhance the agent’s ability to accurately identify key segments and effectively reduce ambiguity and uncertainty in decision-making.

\subsection{Uncertainty Aware Reward Fusion}
\label{sec:Uncertainty-Aware-Reward-Fusion}
During video exploration process, the importance of segments needs to be clearly assessed to effectively guide the exploration direction.
Existing methods~\cite{yang2025vca} typically rely directly on Vision-Language Models as intrinsic reward models; however, these rewards heavily depend on sparsely and uniformly sampled frames.
Due to the inherent randomness of this sampling strategy, the agent may  miss critical information within video segments, causing semantically distinct segments to receive similar rewards.
With the semantic queries discovered and continuously updated during exploration, we introduce the query score, which evaluates the relevance and significance of queries within each segment. Furthermore, we propose Uncertainty Aware Reward Fusion to adaptively integrates intrinsic rewards and semantic query scores, thus enhancing the discriminability of segment evaluations. Specifically, for each candidate segment $s$, the agent computes two scores:

1. \textbf{Intrinsic reward} $r(s)$ — obtained from a VLM through Chain-of-Thought~\cite{wei2022chain} reasoning, reflecting the segment's relevance or informativeness with respect to the query.

2. \textbf{Query score} $u(s)$ — derived by performing softmax pooling on the semantic similarity scores ${\phi_j}$ of all anchors that fall within $s$ denoted as $\mathcal{H}(s)$. Specifically, let the set of similarity scores be ${\phi_j}_{j \in \mathcal{H}(s)}$ and the temperature coefficient be $\tau_c$, then the query score $u(s)$ is computed as:
\[
u(s) = \tau_c \cdot \log \left( \frac{1}{|\mathcal{H}(s)|} \sum_{j \in \mathcal{H}(s)} \exp\left(\frac{\phi_j}{\tau_c}\right) \right)
\]

To quantify the uncertainty of the intrinsic rewards, we compute the normalized entropy $H$ over the reward distribution:
\[
H = -\frac{1}{\log N}\sum_{i=1}^{N}p_i\log p_i,
\]
where $p_i$ denotes the softmax-normalized probability of each candidate segment based on $r(s)$, and $N$ is the total number of candidate segments in the current search state.
A higher $H$ indicates greater uncertainty in the reward distribution, while a lower $H$ implies stronger confidence.

Finally, the intrinsic reward $r(s)$ and query score $u(s)$ are fused in an adaptive manner:
\[
h(s) = (1 - H) \cdot r(s) + H \cdot u(s).
\]
This adaptive weighting allows th
e agent to rely more on intrinsic rewards when the reward distribution is confident (low $H$), and to place greater emphasis on semantic priors when the reward model exhibits high uncertainty.
The fused reward $h(s)$ is subsequently used for exploration decision-making in the selection step.
\subsection{Semantic Guided Tree Search}
\label{sec:tree-search}
In our framework, video exploration is formulated as a tree-search process, where each node represents a temporal segment and the agent progressively expands nodes to locate key evidence.
Unlike previous methods that rely solely on exploration~\cite{yang2025vca,wang2024videotreeadaptivetreebasedvideo} or query-based heuristics~\cite{VideoAgent}, we introduce Semantic Guided Expansion (SGE), which integrates semantic anchors and coverage sampling to achieve efficient and informed exploration.
Specifically, the semantic anchors provide strong guidance by prioritizing semantically relevant regions, while the coverage-driven strategy ensures sufficient exploration of unexplored areas, thereby maintaining a robust balance between exploration and exploitation throughout the search process.

Given the current video segment as a node $s$ and the frame sampling budget $B$, the agent performs frame sampling by combining semantic guidance and temporal coverage. The sampled frames $F$ are then used to divide $s$ into several new child nodes $s_i$.
First, among the semantic anchors contained within node $s$, the agent selects the top $B_s$ frames with the highest similarity scores.
Then, to maximize temporal coverage, the agent selects an additional $B - B_s$ frames on top of the previously chosen $B_s$ frames, forming a final set $F$ of size $B$. The final set $F$ satisfies the following temporal coverage minimax objective:
for any frame $f$ in the video, let $d(f, F)$ denote the minimum temporal distance between any frame $f$ and its nearest frame in $F$.
The remaining frames are chosen such that:
\[
\min_{F} \max_{f \in \text{video}} d(f, F)
\]
is achieved, thereby maximizing temporal coverage with in a segment and reducing long unobserved intervals.
This semantic-guided expansion strategy enables the agent to efficiently locate key information while minimizing the omission of information.

\begin{algorithm}[t]
\small
\caption{\small \Model: Exploitation-Exploration Agent}
\label{algo:sgcee}
\begin{algorithmic}[1]
\REQUIRE Video $V$, user instruction $q$, reward model $R$, policy $\pi$, frame budget $B$, semantic anchor budget $B_s$
\STATE Initialize selected segment and $s^{*} \gets V$
\STATE Initialize candidate segments set $S \gets \emptyset$
\STATE Initialize memory buffer $M \gets \emptyset$
\STATE Initialize reward score history $H_r \gets \emptyset$
\STATE Initialize semantic query set $\mathcal{C} \gets \text{QueryDiscovery}(q)$
\STATE Initialize semantic anchor set $\mathcal{A} \gets \text{InitialAnchors}(\mathcal{C})$

\WHILE{\texttt{true}}
    \STATE \texttt{\textcolor{blue}{\# Sampling \& Expansion}}
    \STATE $\left\{f^{i}_{\text{$\mathcal{A}, s^*$}}\right\}_{i=1}^{B_s} \gets \text{SelectAnchors}(s^{*}, \mathcal{A})$
    \STATE $\left\{f^{i}_{s^{*}}\right\}_{i=1}^{B},\;
           \left\{s^{*}_{i}\right\}_{i=1}^{B+1}
           \gets \text{CoverageSample}(s^{*}, \left\{f^{i}_{\text{$\mathcal{A}, s^{*}$}}\right\})$

    \STATE \texttt{\textcolor{blue}{\# Evaluation}}
    \STATE $\left\{t_{i}, r_{i}\right\}_{i=1}^{B+1}
           \gets R\!\left(\left\{s^{*}_{i}\right\}_{i=1}^{B + 1}, q,
           \left\{f^{i}_{s^{*}}\right\}_{i=1}^{B}, H_r\right)$
    \STATE $\left\{u_{i}\right\}_{i=1}^{B+1}
           \gets \text{QueryScoring}\!\left(\left\{s^{*}_{i}\right\}_{i=1}^{B + 1}, \mathcal{A}\right)$

    \STATE \texttt{\textcolor{blue}{\# Candidate update}}
    \STATE $S \gets \left(S \setminus \left\{s^{*}\right\}\right)
                 \cup \left\{\left(s^{*}_{i}, r_{s_i^{*}}, u_{s_i^{*}}\right)\right\}_{i=1}^{B+1}$

    \STATE \texttt{\textcolor{blue}{\# Uncertainty-Aware Reward Fusion}}
    \STATE $\left\{h_{s}\right\}_{s \in S}
           \gets \text{UARF}\!\left(\left\{(r_{s}, u_{s})\right\}_{s \in S}\right)$

    \STATE \texttt{\textcolor{blue}{\# Logs \& memory}}
    \STATE $H_r \gets H_r \cup \left\{\left(t_{i}, r_{i}\right)\right\}_{i=1}^{B+1}$
    \STATE $M \gets \text{UpdateMemory}\!\left(M, \left\{f^{i}_{s^{*}}\right\}_{i=1}^{B}, \left\{r_{s_i^*}\right\}_{i=1}^{B+1}\right)$

    \STATE \texttt{\textcolor{blue}{\# Query \& anchor update}}
    \STATE $\Delta\mathcal{C} \gets
           \text{QueryUpdate}\!\left(q, \left\{f^{i}_{s^{*}}\right\}\right)$
    \STATE $\mathcal{C},\mathcal{A} \gets \mathcal{C} \cup \Delta\mathcal{C},\text{UpdateAnchors}\!\left(\mathcal{A}, \Delta\mathcal{C}\right)$

    \STATE \texttt{\textcolor{blue}{\# Selection}}
    \STATE $\mathit{ans},\, s_{\text{next}} \gets \pi\!\left(q, {s}_{s \in S}, \left\{h_{s}\right\}, M\right)$
    \IF{$\mathit{ans} \neq \varnothing$}
        \RETURN $\mathit{ans}$
    \ELSE
        \STATE $s^{*} \gets s_{\text{next}}$
    \ENDIF
\ENDWHILE
\end{algorithmic}
\end{algorithm}
The process of semantic guided tree search can be summarized in the following four steps:

\noindent\textbf{(i) Expansion:}  
The agent performs \textit{Semantic-Guided Expansion} on the selected leaf node $s$, obtaining associated frames $F$ and several new child nodes $s_i$.

\noindent\textbf{(ii) Evaluation:}  
For each child node $s_i$, the agent obtains the reasoning trace $t(s_i)$ and intrinsic reward $r(s_i)$ from the reward model, and then computes the fused reward $h(s_i)$ by \textit{Uncertainty Aware Reward Fusion}.

\noindent\textbf{(iii) Selection:}  
The agent makes decisions based on above information.  
The decision may involve:  
(a) expanding the next leaf node, which may include backtracking and branching; or  
(b) directly answering the query when evidence is sufficient.

\noindent\textbf{(iv) Update:}  
The agent updates both the semantic queries $\mathcal{C}$ and the semantic anchors $\mathcal{A}$ as \textit{Dynamic Query Management}.  
The associated frames $F$ are added to the memory buffer, and low reward frames are removed when the buffer exceeds its capacity.

\begin{table*}[t]
    \begin{minipage}[t]{0.65\textwidth}
        \centering
        \small 
        \begin{tabular}{@{}lcrrrrrrr@{}}
            \toprule
            Method & Frames & ER & EU & KIR & TG & Rea & Sum & Avg.\\
            \midrule
            Gemini~1.5~Pro~\cite{team2024gemini} & 3600 & 32.1 & 30.9 & 39.3 & 31.8 & 27.0 & 32.8 & 33.1 \\
            Gemini-2.0-Flash~\cite{google2024geminiupdate} & 3600 & 47.4 & \underline{48.5} & \textbf{56.8} & 39.3 & 44.4 & 41.4 & 48.6 \\
            TimeChat~\cite{ren2024timechat} & $>$96 & 21.9 & 21.7 & 25.9 & 22.7 & 25.0 & 24.1 & 22.3 \\
            MovieChat~\cite{song2024moviechat} & $>$10000 & 21.3 & 23.1 & 25.9 & 22.3 & 24.0 & 17.2 & 22.5 \\
            InternVL2.5-72B~\cite{chen2024expanding} & 16 & 43.8 & 42.0 & 42.1 & 36.8 & \underline{51.0} & 37.9 & 43.6 \\
            Qwen2.5VL-72B~\cite{bai2025qwen2} & ~768 & - & - & - & - & - & - & 47.3 \\
            GLM4V-Plus~\cite{hong2024cogvlm2visuallanguagemodels} & $\leq$300 & 46.2 & 47.8 & \underline{54.1} & 42.7 & 46.5 & 37.9 & \underline{48.7} \\
            \midrule
            GPT-4o~\cite{hurst2024gpt} & 64 & 35.9 & 30.8 & 35.5 & 28.3 & 33.5 & 34.5 & 34.7 \\
            VideoAgent~\cite{VideoAgent} & Avg. 25.5 & 28.0 & 30.3 & 28.0 & 29.3 & 28.0 & 36.4 & 29.3 \\
            VideoTree~\cite{wang2024videotreeadaptivetreebasedvideo} & Avg. 103.2 & 30.3 & 25.1 & 26.5 & 27.7 & 31.9 & 25.5 & 28.8 \\
            VCA~\cite{yang2025vca} & Avg. 20.0 & 43.7 & 40.7 & 37.8 & 38.0 & 46.2 & 27.3 & 41.3 \\
            \Model (ours) & \textbf{Avg. 13.5} & \textbf{52.9} & \textbf{54.0} & 53.0 & \textbf{58.7} & \textbf{55.4} & \underline{47.3} & \textbf{53.6} \\
            \midrule
            Seed1.6VL~\cite{guo2025seed1} & 64 & 30.7 & 29.8 & 33.8 & 27.7 & 33.6 & 30.6 & 31.5 \\
            VCA~\cite{yang2025vca} & Avg. 21.2 & 39.5 & 36.9 & 38.4 & 40.5 & 35.7 & 32.4 & 38.7 \\
            \Model (ours) & \underline{Avg. 14.2} & \underline{49.8} & 48.2 & 52.8 & \underline{55.5} & 48.3 & \textbf{50.0} & \underline{50.8} \\
            \bottomrule
        \end{tabular}
        \vspace{-0.5em}
        \caption{Experimental Results on LVBench. For agent-based methods, the average number of frames observed is reported.}
        \vspace{-1.8em}
        \label{exp:lvbench-main}
    \end{minipage}
    \hfill 
    \begin{minipage}[t]{0.3\textwidth}
        \centering
        \small 
        \captionsetup{width=\textwidth} 
        \begin{tabular}{@{}lcr@{}}
            \toprule
            Method & Frames & Subset \\
            \midrule
            {GPT-4o~\cite{hurst2024gpt}} & 32 & 70.4 \\
            \multirow{2}{*}{VideoAgent~\cite{VideoAgent}} & Avg. 8.1 & 63.2 \\
             & Avg. 10.7 & 60.8 \\
            VideoTree~\cite{wang2024videotreeadaptivetreebasedvideo} & Avg. 69.5 & 67.0 \\
             LVNet~\cite{park2024too} & 12 & 68.2  \\
            VCA~\cite{yang2025vca} & \underline{Avg. 7.2} & 73.6 \\
            \Model (ours) & \textbf{Avg. 6.1} & 75.6 \\
            \midrule
            Seed1.6VL~\cite{guo2025seed1} & 32 & 72.4 \\
            VCA~\cite{yang2025vca} & Avg. 8.6 & 74.1  \\
            \Model (ours) & Avg. 6.9 & \underline{76.3} \\
            \midrule
            {Qwen2-VL}~\cite{Qwen2VL} 
            & 32 & 74.2 \\
            VideoAgent~\cite{VideoAgent} & Avg. 8.6 & 65.2 \\
            VideoTree~\cite{wang2024videotreeadaptivetreebasedvideo} & Avg. 85.4 & 71.0\\
            VCA~\cite{yang2025vca}  & Avg. 16.9 & 75.2 \\
            \Model  & Avg. 12.6 & \textbf{77.9} \\
            \bottomrule
        \end{tabular}
        \vspace{-0.5em}
        \caption{{Experimental Results on EgoSchema}.}
        \vspace{-1.8em}
        \label{exp:egoschema-main}
    \end{minipage}
\end{table*}

The overall framework of the proposed method is outlined in Algorithm~\ref{algo:sgcee}.
In summary, our semantic guided tree search enables the agent to emulate the human cognitive process through an iterative cycle of exploring, integrating new queries, and refining understanding.
Moreover, the hierarchical design and semantic-guided exploration make the framework scalable with respect to video length, maintaining efficient and reliable performance as the video duration increases.
\section{Experiments}
\subsection{Experimental Settings}
\label{sec:exp-setup}
 \textbf{Benchmark.}  We mainly evaluate \Model on two widely used long-video QA benchmarks. \textbf{EgoSchema}~\cite{mangalam2023egoschema} is built from Ego4D~\cite{grauman2022ego4d} and uses 3-minute egocentric clips with 5-way multiple-choice questions; following common practice, we report results on the official validation subset of 500 questions. 
\textbf{LVBench}~\cite{wang2024lvbench} targets \emph{extreme} long videos and stresses long-term memory with diverse categories (e.g., TV series, sports, surveillance) where videos frequently exceed 30 minutes; we follow the official evaluation split released by the authors.
We also conduct experiments on other relevant benchmark datasets in Appendix.
We further evaluate \Model on additional related datasets, with detailed results provided in the Appendix. 

\textbf{Baselines.} 
We conduct comparisons with the most relevant and competitive baselines.
These include leading open-source Video-LLMs~\cite{ ren2024timechat,song2024moviechat,chen2024expanding,bai2025qwen2,hong2024cogvlm2visuallanguagemodels}, the proprietary large multimodal model GPT-4o~\cite{hurst2024gpt}, Seed-VL~\cite{guo2025seed1} Gemini model~\cite{team2024gemini,google2024geminiupdate}, and a set of representative agent-based approaches~\cite{VideoAgent, wang2024videotreeadaptivetreebasedvideo, yang2025vca, park2024too}.
For video models, we report results obtained using uniformly sampled frames.
For agent-based systems, we adopt their official experimental settings to ensure a fair comparison. 

 \textbf{Implementation Details.} We use a single model for both reward and policy model. 
For fair comparison with baseline methods, we conduct experiments using GPT-4o-2024-0806.
To further verify that our approach is model-independent, we also perform experiments on Seed-1.6-250615~\cite{guo2025seed1}.
When invoking LLM API services, we set the temperature to 0.5.
The memory buffer sizes are set to 8 and 16 for EgoSchema and LVBench.
The temperature coefficient $\tau_c$ for query score is 0.1.

\subsection{Experimental Results}
We conduct comparative experiments on two challenging long-video understanding benchmarks,  {LVBench} and  {EgoSchema}, with the results presented in Tab.~\ref{exp:lvbench-main} and Tab.~\ref{exp:egoschema-main}, respectively. 
By explicitly integrating both exploration and exploitation behaviors, our agent achieves not only substantial performance gains but also significant improvements in computational efficiency. 
When utilizing only around 20\% of the video frames, \Model yields clear advantages over directly feeding uniformly sampled frames into large multimodal models. 
Compared with  {GPT-4o}, \Model achieves improvements of \textbf{+18.9\%} on LVBench and \textbf{+5.2\%} on EgoSchema; compared with {Seed1.6VL}, the gains are \textbf{+19.3\%} and \textbf{+3.9\%}, respectively. 
Notably, despite operating with less than \textbf{0.5\%} of the frame sampling budget, our method still surpasses the state-of-the-art closed-source model  {Gemini-2.0-Flash} by over \textbf{5\%}, highlighting its remarkable efficiency and capability to focus on truly informative segments.

Furthermore, we compare our proposed framework with three recently introduced state-of-the-art on-the-fly agent systems, including  {VideoAgent}~\cite{VideoAgent},  {VideoTree}~\cite{wang2024videotreeadaptivetreebasedvideo}, and  {VCA}~\cite{yang2025vca}. 
As shown in Tab.~\ref{exp:lvbench-main}, \Model achieves more than \textbf{10\%} higher accuracy on  {LVBench} compared to all baseline methods, while processing at least \textbf{30\%} fewer frames. 
This indicates that relying solely on semantic priors (as in VideoAgent) or discovering information through uniform sampling without any prior knowledge (as in VideoTree and VCA) is highly inefficient. 
In contrast, our method integrates semantic prior–driven exploitation with prior-free exploration that maximizes coverage of unseen regions, thereby substantially improving both the efficiency and success rate of discovering key information. 
As a result, \Model shows remarkable improvements on the \emph{Key Information Retrieval} task of  {LVBench} compared with other agent-based systems. 
Meanwhile, the results on  {EgoSchema} also demonstrate the significant advantages of our approach in both performance and computational efficiency. 
Additional experimental results are provided in Appendix.

\Model exhibits not only strong performance when built upon advanced proprietary backbones such as GPT-4o and Seed1.6VL, but also clear advantages when applied to open-source models.
For a fair comparison, we conduct experiments using {Qwen2-VL-72B-Instruct-GPTQ-Int4}~\cite{Qwen2VL} as the underlying backbone.
As shown in Tab.~\ref{exp:egoschema-main}, our agent achieves a \textbf{3.7\%} improvement in accuracy on the EgoSchema benchmark while observing less than \textbf{40\%} of the frames compared to directly feeding uniformly sampled frames into Qwen2-VL.
Moreover, relative to other agent-based systems, \Model attains an additional \textbf{2.7\%} performance gain while using at least \textbf{25\%} fewer frames.
These results confirm that the proposed framework {retains its effectiveness and generalization capability beyond proprietary architectures}.

\begin{table}[!t]
    \centering
    \small
    \scalebox{1}{
    \begin{tabular}{@{}lcrcr@{}}
        \toprule
        \multirow{2}{*}{Method} & \multicolumn{2}{c}{EgoSchema} & \multicolumn{2}{c}{LVBench} \\
        & Frames & Acc. & Frames & Acc. \\
        \midrule
        \Model (ours) & 6.9 & \textbf{76.3} & 14.2 & \textbf{50.8} \\
        - w/o SGE & 10.6 & 72.6 & 21.3 & 42.5 \\
        - w/o UARF & 9.2 & 73.2 & 18.0 & 47.8 \\
        - w/o QU & 7.5 & \underline{75.5} & 15.4 & \underline{48.6} \\
        \bottomrule
    \end{tabular}
    }
    \caption{{Ablation Study on EgoSchema and LVBench}. 
    SGE denotes semantic-guided expansion, UARF denotes uncertainty-aware reward fusion, and QU denotes query update.
    }
    \label{exp:ablation-egoschema-lvbench}
\end{table}

\begin{figure}[tbp]
\centering
\centerline{\includegraphics[width=0.85\linewidth]{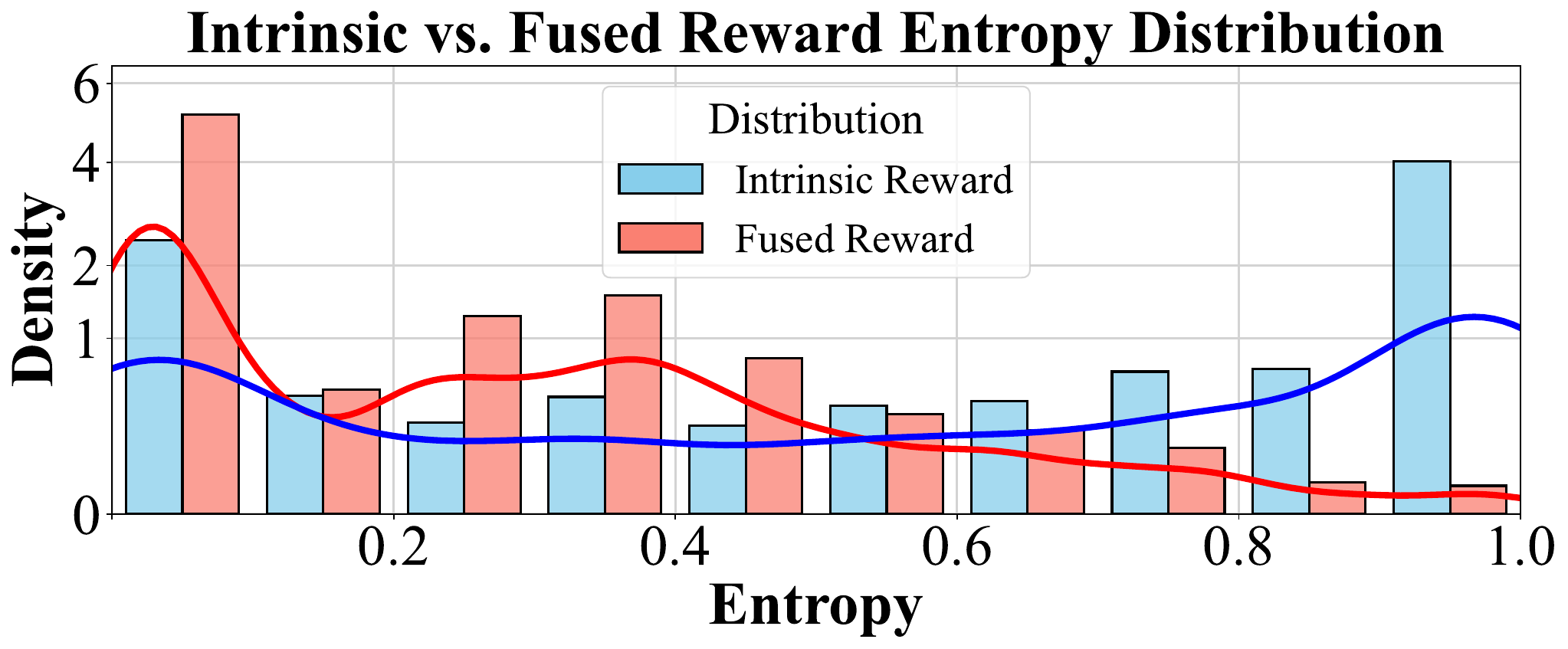}}
\caption{\textbf{Comparison of Entropy Distribution between Intrinsic Reward and Fused Reward on LVBench.} The y-axis represents the value of the entropy, and the x-axis represents the probability density. The red 
and blue curves represent fitted normal distributions.}
\label{fig:reward-match}
\end{figure}

\begin{figure*}[htbp]
\centering
\centerline{\includegraphics[width=\linewidth]{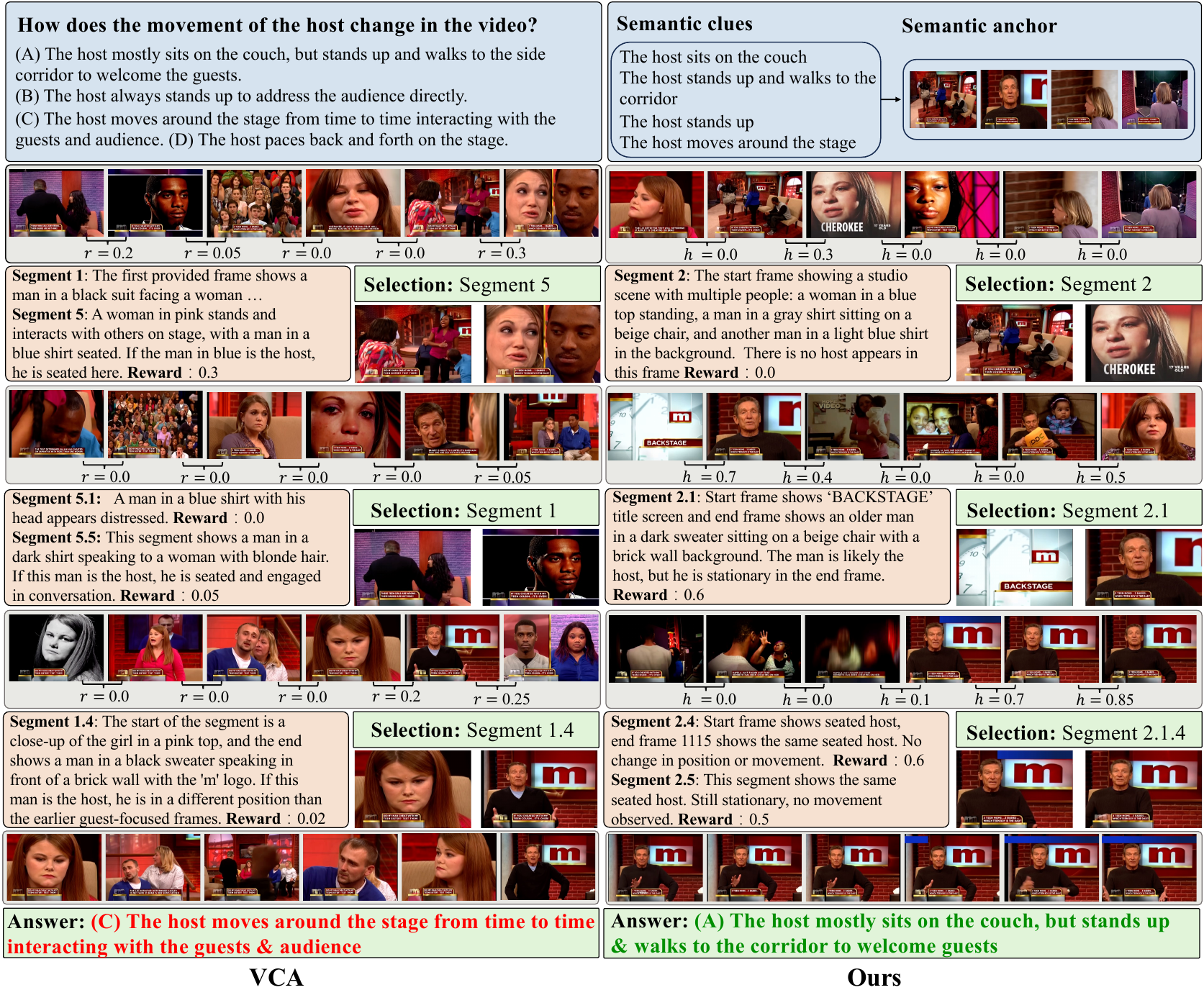}}
\vspace{-0.5em}
\caption{\textbf{Exploration Trajectory Example.} Compared with VCA, our agent can rapidly pinpoint critical information in each segment with the guidance of semantic queries and anchors. Moreover, even when the reward model becomes unreliable, the agent can still produce a discriminative fused reward by integrating the query score.}
\vspace{-1.5em}
\label{fig:good-case}
\end{figure*}

\subsection{Ablation Study}
In this section, we conduct ablation studies on the key components of our agent, with the results summarized in Tab.~3. 
As shown in the table, removing the {semantic-guided expansion} leads to a noticeable performance drop, accompanied by a significant increase in the number of frames required for decision-making. 
This occurs because, without semantic priors, the model relies on uniform sampling to explore unknown segments and often misses critical visual information, thereby degrading its understanding of video content. 
Consequently, the agent tends to examine many irrelevant frames, which substantially reduces computational efficiency. 
These findings highlight the effectiveness of incorporating semantic priors during the node expansion process in tree search.

In addition, we observe a substantial performance decline when the {uncertainty-aware reward fusion} is removed. 
In this case, the agent relies solely on intrinsic rewards obtained from the VLM to evaluate the importance of each segment. 
However, during a round of tree search, the model use sparsely sampled frames to represents each segment, which may not fall within regions containing crucial information—resulting in low discriminability of the rewards. 
As a result, relying solely on intrinsic rewards prevents the agent from making reliable decisions and also contributes to reduced computational efficiency. 
These results demonstrate the necessity of integrating the {query score}, which introduces semantic priors into the evaluation and decision-making process.

We also find that removing the {query update} mechanism causes a clear degradation in performance. 
This is because the initial clues discovered from the query are often incomplete, and as the agent continues to explore, it observes new visual evidence and discovers additional semantic queries. 
These newly discovered clues, in turn, help the agent identify more critical content during the {semantic-guided expansion} stage and yield more reliable {query scores} during the evaluation stage, forming a positive feedback loop that enhances overall reasoning performance.

\subsection{How Effective Is the Reward Fusion?}
To further analyze the effectiveness of our uncertainty-aware reward fusion mechanism, we compare the entropy distributions obtained from the original intrinsic reward and the fused reward on LVBench, as shown in Fig.~\ref{fig:reward-match}.
This experiment examines whether the fusion mechanism can address the low discriminability of the intrinsic reward, thereby providing the agent with a more stable evaluation signal.
From Fig.~\ref{fig:reward-match}, we observe that the entropy distribution of the intrinsic reward exhibits a clear bimodal pattern, with high-entropy and low-entropy regions dominating most cases.
High entropy with high density indicates that the intrinsic reward often lacks discriminative power.
In contrast, the fused reward distribution shows a substantial reduction in probability density within the high-entropy region. This demonstrates that when the intrinsic reward is highly uncertain, the query-based score effectively helps separate different segments through UARF, resulting in a more informative reward.
Overall, these results indicate that our uncertainty-aware reward fusion strategy enhances reward discriminability under high uncertainty scenarios, enabling the agent to obtain reliable reward signals during exploration.

\subsection{How Does \Model Explore?}
We compare the real exploration trajectories of the baseline method {VCA} and our \Model on {LVBench}, as illustrated in Fig.~\ref{fig:good-case}. 
First, with the help of {semantic-guided expansion}, our model is more likely to focus on information relevant to user query when expanding video segments. 
In contrast, {VCA}, which conducts uniform sampling in expansion, tends to collect segments that are mostly irrelevant to the question; throughout the exploration process, it receives only a limited amount of visual evidence related to the host, which ultimately leads to an incorrect answer. 
Second, as shown in the figure, during the first expansion round, the reward model assigns an incorrect score of zero to the key segment. 
However, since our method evaluates segments by combining the {query score} with the intrinsic reward, it successfully gives the correct reward and identifies the segment containing rich host-related information.


\section{Conclusion}
In this work, we propose \Model a long video agent framework that balances exploration and exploitation through semantic guidance. 
The agent extracts and continually updates semantic queries, incorporates them into semantic guided expansion and uncertainty-aware reward fusion in hierarchical tree search, and achieves significantly improved performance and efficiency.
In the future, we plan to further improve the scalability of our framework to handle longer and more complex videos, and explore efficient
mechanisms for real-time and continuous understanding of dynamic video streams.

{
    \small
    \bibliographystyle{ieeenat_fullname}
    \bibliography{main}
}

\clearpage
\setcounter{page}{1}
\maketitlesupplementary

\section{Implementation Details}
\subsection{Dataset}
In this section, we provide detailed descriptions and statistics of the benchmark datasets. \\
\textbf{EgoSchema}~\cite{mangalam2023egoschema} is built from Ego4D~\cite{grauman2022ego4d} and uses 3-minute egocentric clips with 5-way multiple-choice questions; following common practice, we report results on the official validation subset, which contains 500 distinct videos, each associated with a single question. \\
\textbf{LVBench}~\cite{wang2024lvbench} targets \emph{extreme} long videos and stresses long-term memory with diverse categories (e.g., TV series, sports, surveillance) where videos frequently exceed 30 minutes. Due to seven videos becoming unavailable for download (28CIeC8cZks, QWXlvx1GoTY, QgWRyDV9Ozs, gXnhqF0TqqI, idZkam9zqAs, qYMnM5blZIE, t-RtDI2RWQs), we use a total of 96 videos in our experiments. These videos have an average duration of 3,942 seconds and correspond to 1,432 questions in total. \\
\textbf{Video-MME}~\cite{fu2024vide} consists of 900 manually selected videos across diverse real-world sources and scen    arios, covering six domains and thirty subcategories, with each video annotated with three expert-verified QA pairs. In this work, we focus on the subset of videos categorized as long in duration to assess models’ long-video understanding capability. This subset contains 300 videos with an average length of 2,466 seconds and a total of 900 associated questions.\\
\textbf{MMBench-video}~\cite{fang2024mmbenchvideo} is a comprehensive VideoQA benchmark built from diverse YouTube videos, designed to evaluate LVLMs across 26 capability dimensions, including perception, spatial–temporal reasoning, multimodal grounding, and high-level semantic reasoning. The dataset contains 609 videos with an average duration of 165 seconds, accompanied by 1,998 free-form QA pairs. To ensure fair comparison with previous work~\cite{yang2025vca}, we adopt GPT-4o as the evaluator.

While other long-video benchmarks are available, many of them are built from movie sources. Examples include MovieChat~\cite{song2023moviechat}, MovieQA~\cite{tapaswi2016movieqa}, and related movie-based datasets~\cite{zhang2023movqa,soldan2022mad}. Such datasets carry a potential risk of overlapping with the pre-training corpora of recent LVLMs, such as GPT-4o. To ensure a fair and uncontaminated evaluation, we therefore refrain from using these movie-based benchmarks. This practice is consistent with previous work~\cite{yang2025vca}, which similarly excludes movie-derived datasets to prevent unintended data overlap with proprietary LVLMs.

\subsection{Baselines}
In this section, we present the implementation details of all baseline methods.
For a fair comparison across all baselines, we ensure that all methods are evaluated using the same backbone LLM (GPT-4o).
Because the original VideoAgent~\cite{VideoAgent} and Videotree~\cite{wang2024videotreeadaptivetreebasedvideo} implementation relied on GPT-4, while our evaluation uses GPT-4o, we adopt the VideoAgent and Videotree results reproduced by VCA~\cite{yang2025vca}, which evaluated the method under the same GPT-4o setting as ours.

Since VCA has not released its source code, we reimplemented the method based on the algorithmic descriptions and prompt templates provided in the original paper. We report the results using Seed1.6VL as both the reward and policy model in the main text. For a fair comparison, we follow the same hyperparameter settings as our method: on LVBench, we set the sampling frame number $N=6$ and the memory buffer size to 16; on EgoSchema, we use $N=4$ and a memory buffer size of~8.

\begin{table*}[htbp]
    \centering
    \begin{tabular}{@{}lrrrr@{}}
        \toprule
        Method & VideoAgent & VideoTree & VCA & Ours \\
        \midrule
        Avg. Frames & 24.6 & 98.0 & \underline{18.1} & \textbf{14.2} \\
        \midrule
        Knowledge & 52.2 & \underline{60.7} & 56.9 & \textbf{62.3} \\
        Film \& Television & 42.5 & 52.5 & \underline{55.0} & \textbf{56.9} \\
        Sports Competition & 42.7 & 48.6 & \underline{59.3} & \textbf{65.0} \\
        Artistic Performance & 47.5 & 51.6 & \textbf{65.8} & \underline{63.3} \\
        Life Record & 44.7 & 49.5 & \textbf{51.9} & \underline{50.4} \\
        Multilingual & 36.6 & 40.0 & \underline{46.7} & \textbf{52.1} \\
        \midrule
        Overall & 46.4 & 53.1 & \underline{56.3} & \textbf{59.0} \\
        \bottomrule
    \end{tabular}
    \caption{\textbf{Experimental Results on VideoMME Long Split}. We list the average observed frames inspected by each method, marking the best performance in \textbf{bold} and the second best with \underline{underline}.}
    \label{exp:videomme-main}
\end{table*}

\begin{table*}[htbp]
    \centering
    \scalebox{0.9}{
    \begin{tabular}{@{}lrrrrr@{}}
        \toprule
        Method & GPT-4o & VideoAgent & VideoTree & VCA & Ours \\
        \midrule
        Avg. Frames & 8 & 7.8 & 27.1 & \underline{7.4} & \textbf{6.1} \\
        Score & 1.62 & 1.05 & 1.38 & \underline{1.68} & \textbf{1.93} \\
        \bottomrule
    \end{tabular}
    }
    \caption{\textbf{Experimental Results on MMBench-Video}. We list the average observed frames inspected by each method, marking the best performance in \textbf{bold} and the second best with \underline{underline}.}
    \label{exp:mmbench-main}
\end{table*}

\section{Additional Quantitative Results}

\subsection{Additional Comparasion Results}
The experimental results of our method and the baseline methods on VideoMME and MMBench-Video are presented in Table~\ref{exp:videomme-main} and Table~\ref{exp:mmbench-main}, respectively.
The experimental results demonstrate that our method achieves substantial improvements in both performance and efficiency.
As shown in Table~\ref{exp:videomme-main}, \Model\ surpasses VCA by 2.7\% in accuracy while reducing the number of observed frames by more than 20\%; it also significantly outperforms VideoAgent and VideoTree by 12.6\% and 5.9\%, respectively.
In Table~\ref{exp:mmbench-main}, we observe that, while using far fewer frames, our method attains performance close to that of GPT-4o and achieves a notable gain over VCA. We attribute this improvement to the ability of our framework to more efficiently discover fine-grained visual clues.

\subsection{Additional Ablation Study}

\begin{table*}[h!]
    \centering
    \begin{tabular}{@{}lrrrrrrrr@{}}
        \toprule
        Method & Avg. Frames & ER & EU & KIR & TG & Rea & Sum & Avg. \\
        \midrule
        Ours        & 14.2 & \textbf{49.8} & \textbf{48.2} & \textbf{52.8} & \textbf{55.5} & \textbf{48.3} & \textbf{50.0} & \textbf{50.8} \\
        \midrule
        - w/o SGE   & 21.3 & 39.9 & 38.9 & 45.2 & 47.3 & 45.6 & 42.3 & 42.5 \\
        - w/o UARF  & 18.0 & 49.0 & 45.9 & 51.1 & 49.3 & 43.7 & 48.2 & 47.8 \\
        - w/o QU    & 15.4  & 47.5 & 46.2 & 50.2 & 52.2 & 47.1 & 45.5 & 48.6 \\
        \bottomrule
    \end{tabular}
    \caption{\textbf{Ablation Study on LVBench}. We report accuracy across six reasoning categories (ER, EU, KIR, TG, Rea, Sum) and the overall average. Best results per column are highlighted in \textbf{bold}.}
    \label{exp-ablation-lvbench-3variants}
\end{table*}

\textbf{Detailed Ablation Study Result on LVBench.} As shown in Table~\ref{exp-ablation-lvbench-3variants}, removing any core component leads to a performance drop, indicating that the three modules play complementary roles in long-video reasoning. Among them, removing Semantic-Guided Expansion(SGE) results in the largest degradation, with the average accuracy decreasing from 50.8\% to 42.5\% , demonstrating that SGE is crucial for directing exploration toward informative video segments. Removing Uncertainty-Aware Reward Fusion(UARF) causes declines across several reasoning-related dimensions (e.g., \textit{Event Understanding} and \textit{Reasoning}), highlighting its importance in stabilizing reward signals and improving the discriminability of key visual cues. Removing QU primarily affects \textit{Temporal Grounding} and \textit{Summarization}, suggesting that iteratively updating semantic anchors is essential for uncovering previously unobserved but relevant information. Overall, the full model achieves the best balance between accuracy and frame efficiency, attaining the highest average accuracy (50.8\%) while observing fewer frames than all ablated variants.

\begin{figure*}[t]
    \centering
    \centerline{
    \includegraphics[width=0.7\textwidth]{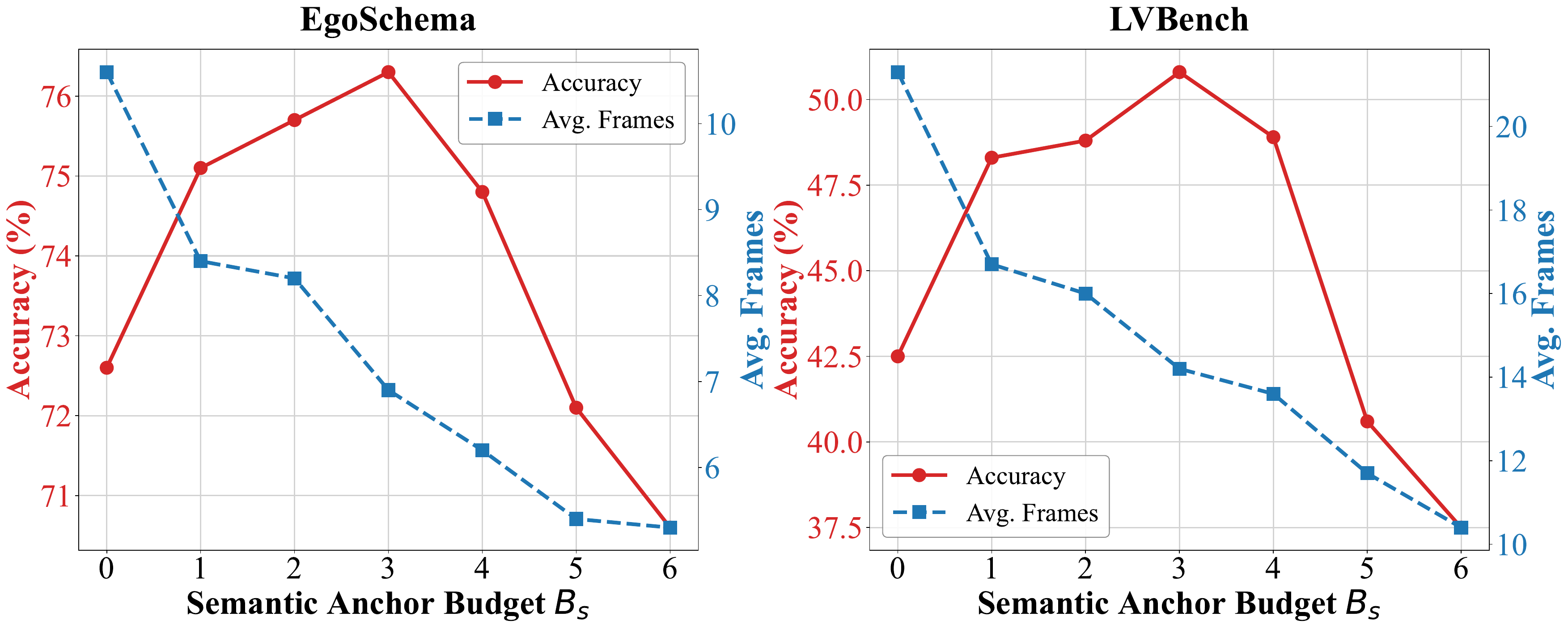}}
\caption{\textbf{Impact of the semantic anchor budget $B_s$.} 
Accuracy (red solid line) and average observed frames (blue dashed line) on EgoSchema and LVBench. 
Both datasets exhibit a clear peak at $B_s=3$, with performance dropping when the anchor budget is too small or too large, 
indicating that a moderate semantic budget achieves the best balance between exploration quality and frame efficiency.}
    \label{fig:b_s}

\end{figure*}

\textbf{Impact of the semantic anchor budget $\mathbf{B_s}$.} 
As shown in Figure~\ref{fig:b_s}, the choice of $B_s$ plays a critical role in balancing reasoning accuracy and computational efficiency. 
Across both EgoSchema and LVBench, we observe a trend in which accuracy first increases and then decreases as $B_s$ grows. 
Specifically, increasing $B_s$ from 0 to 3 yields steady performance improvements, indicating that a moderate number of anchors effectively guides the model toward semantically informative regions. 
When $B_s = 3$, the performance reaches its peak, reflecting an optimal balance between removing redundancy and preserving essential contextual information. 
However, further increasing $B_s$ leads to noticeable performance degradation, suggesting that over-aggressive anchoring restricts the exploration process and causes the model to focus too narrowly on the content suggested by the semantic anchors.

It is worth noting that although larger values of $B_s$ consistently reduce the average number of observed frames, the accuracy declines accordingly. 
This phenomenon arises because, when too many anchors are used, the highest-scoring semantic anchors often exhibit highly similar visual content, which does not necessarily correspond to the segments required to answer the question, thereby weakening the effectiveness of exploration.

\section{Further Analysis}

\begin{figure*}[t]
    \centering
    \centerline{
    \includegraphics[width=0.7\textwidth]{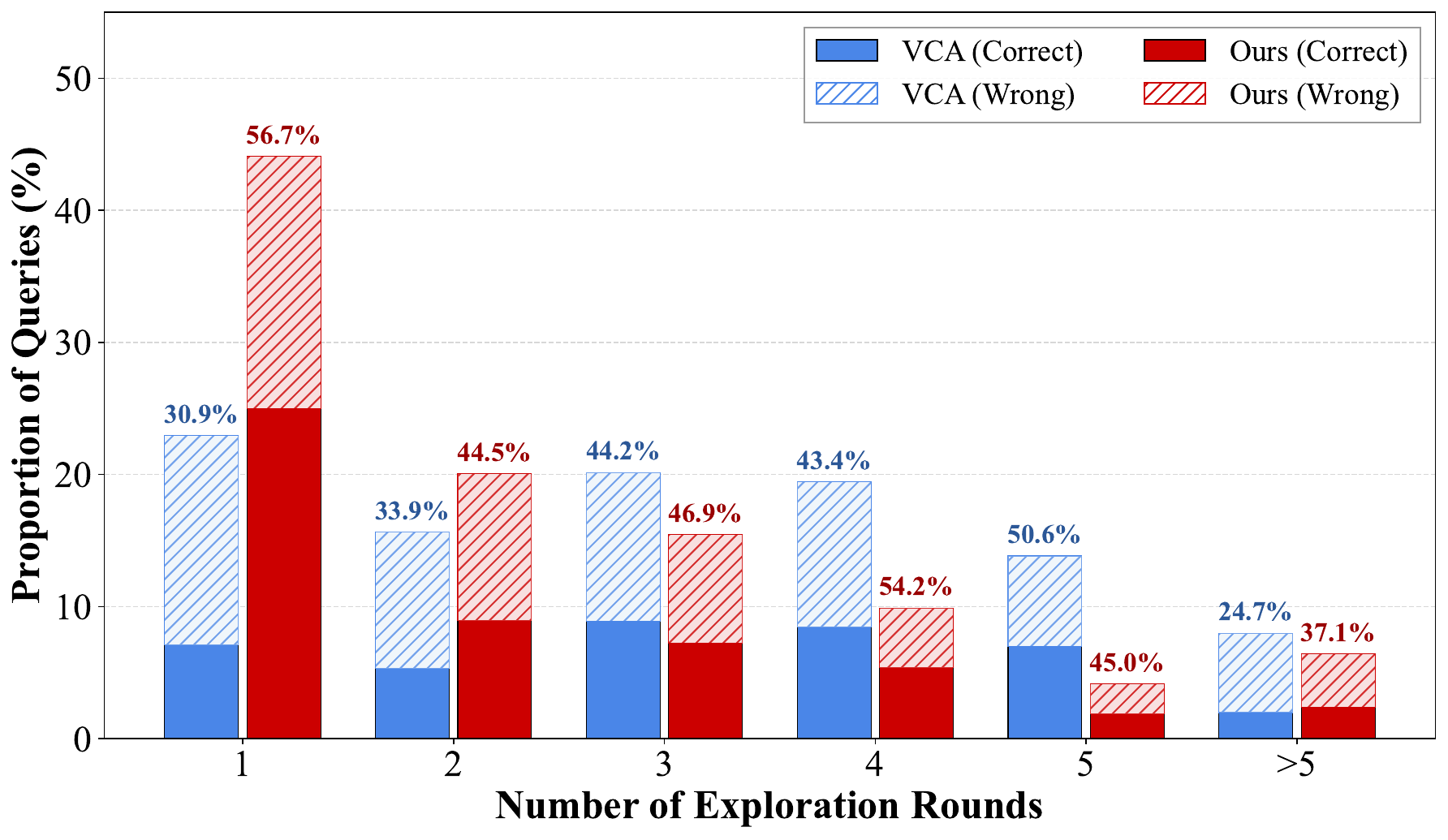}}

\caption{\textbf{Exploration-round statistics.} 
\Model resolves more questions in early rounds and maintains higher accuracy across all depths compared with VCA.}
    \label{fig:round}

\end{figure*}
\subsection{Exploration Efficiency and Depth Analysis.}
Figure~\ref{fig:round} summarizes how questions are resolved across exploration rounds. 
The bar height indicates the termination proportion, and the solid region shows accuracy.

\noindent\textbf{(1) Early-round efficiency.}
\Model resolves a substantially higher fraction of questions in the first round (44.1\% vs.\ 23.0\%; 
56.7\% vs.\ 30.9\% accuracy). 
This front-loaded distribution reflects the benefit of Semantic Guided Expansion (SGE), which directs sampling toward semantically relevant regions, whereas uniform sampling in VCA often misses key evidence.

\noindent\textbf{(2) Semantic refinement in mid exploration rounds.}
In Rounds 2--4, VCA has a larger number of questions that require continued exploration, 
whereas \Model has already resolved many easy cases in the first round and thus handles a more challenging subset in this stage. 
Despite this, \Model still achieves higher accuracy than VCA. 
This advantage comes from Dynamic Query Management (DQM), which continuously refines the semantic scope based on newly observed evidence, keeping the search aligned with relevant content.

\noindent\textbf{(3) Robust deep exploration.}
For questions requiring more than five rounds, \Model maintains a clear accuracy margin (37.1\% vs.\ 24.7\%). 
Through Uncertainty-Aware Reward Fusion (UARF), the agent stabilizes segment evaluation when intrinsic rewards become unreliable, enabling effective reasoning in long-horizon cases.

\noindent\textbf{Overall}, \Model solves more questions at shallow depths and preserves stable performance in deeper rounds, demonstrating consistent advantages over VCA across the entire exploration process.

\begin{figure*}[t]
    \centering
    \centerline{
    \includegraphics[width=0.7\textwidth]{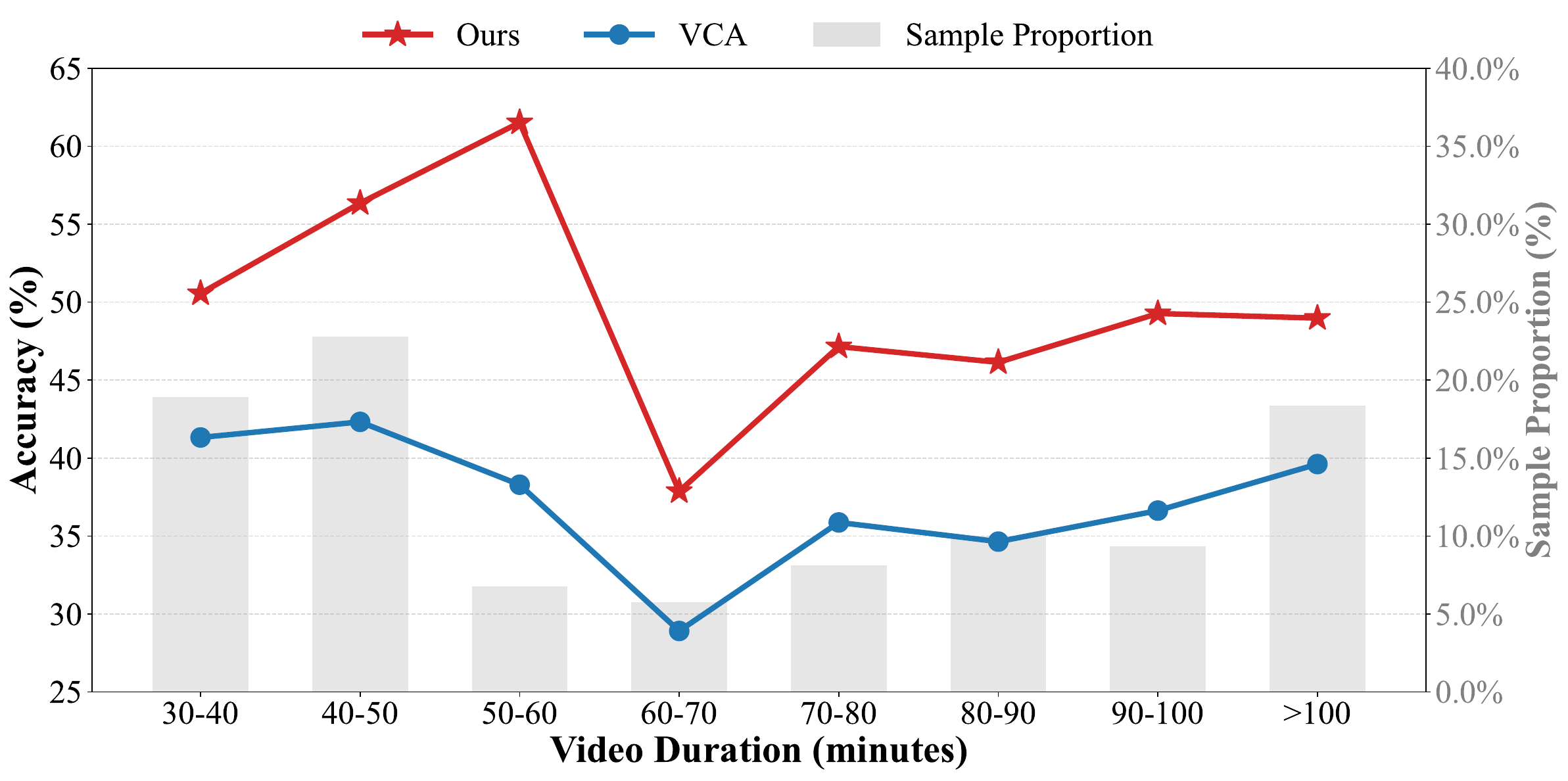}}

\caption{\textbf{Performance across video durations.} 
Accuracy of \Model and VCA on videos grouped by duration. 
\Model consistently outperforms VCA in all ranges, with the largest gains on ultra-long videos where information is sparse and uniform sampling struggles to capture key evidence.}

    \label{fig:duration}

\end{figure*}
\subsection{Robustness Across Video Durations.}
Figure~\ref{fig:duration} shows that \Model consistently outperforms VCA across all duration ranges, with the advantage becoming most pronounced on ultra-long videos (over 100 minutes). This trend aligns well with the overall design of our method. In extremely long temporal windows, informative moments are typically sparse, and uniform sampling provides insufficient coverage of semantically dense regions, making methods relying on such strategies more likely to miss critical evidence.

Long videos present much broader temporal ranges and substantially lower density of informative cues. In this setting, SGE helps avoid searching uniformly across the entire duration by focusing on semantically promising regions, while DQM ensures that the search scope progressively adapts to newly revealed content. These mechanisms reduce unnecessary exploration that becomes especially costly as video length grows. When evidence becomes extremely sparse in ultra-long videos, intrinsic rewards often struggle to distinguish informative segments. UARF mitigates this issue by leveraging semantic priors to stabilize scoring. As a result, \Model maintains reliable localization performance even on multi-hour videos, whereas uniform-sampling methods exhibit significant degradation as duration increases.

\section{Prompts Design}

This section provides a detailed description of the prompt design used in the \Model framework.
Tabs.~\ref{supp:tab_reward_first} and \ref{supp:tab_reward_following} present the prompts for generating intrinsic rewards in the evaluation stage, while Tab.~\ref{supp:tab_policy} shows the prompt used in the selection stage.
Tab.~\ref{supp:tab_query_generation} provides the prompt for the Query Discovery stage, and Tab.~\ref{supp:tab_query_update} presents the prompt for the Query Update stage.

In both the evaluation and selection stages, our prompt design is inspired by the approaches used in VideoTree~\cite{wang2024videotreeadaptivetreebasedvideo} and VCA~\cite{yang2025vca}. Tab.~\ref{supp:tab_reward_first} illustrates how the agent obtains intrinsic rewards during the first round of the evaluation stage, while Tab.~\ref{supp:tab_reward_following} presents the procedure for subsequent rounds, where the key difference is that the agent additionally incorporates the historical relevance scores assigned to other segments. Furthermore, Tab.~\ref{supp:tab_policy} shows how the model leverages the fused reward—obtained by combining intrinsic reward and query score—together with the visual evidence stored in the memory buffer to assess whether sufficient information has been gathered; if not, the agent continues exploration, and if so, it directly outputs the final answer.

In Tab.~\ref{supp:tab_query_generation}, the agent extracts concise and precise semantic queries from the user instructions, including the question and answer options, to obtain semantic anchors. In Tab.~\ref{supp:tab_query_update}, the agent further derives new semantic queries from the newly sampled frames, ensuring that they do not duplicate any existing queries.

\begin{table*}[htbp]
\centering
\begin{minipage}{1\textwidth}\vspace{3mm} 
    \centering
    \caption{\textbf{Prompt for the reward model for the first round, with detailed placeholder descriptions.}}
    \label{supp:tab_reward_first}
    \begin{tcolorbox} 
        \centering
        \hspace{-6mm}
        \begin{tabular}{p{0.99\textwidth}}
        \hspace{1mm} 
        \begin{minipage}{0.99\textwidth}
        \textbf{User} \\
        \textbf{/* Task Description */} \\
        You are acting as a reward model to guide the video question-answering process, with access to a \textcolor{blue}{\texttt{duration}}-frame video (\textcolor{blue}{\texttt{duration}} seconds in duration). You are provided with \textcolor{blue}{\texttt{frame\_number}} uniformly sampled frames from the video, at the following frame indices: \textcolor{blue}{\texttt{frame\_block}}, which divide the video into \textcolor{blue}{\texttt{segment\_number}} distinct segments. \\
        
        \textbf{/* Segment Information */} \\
        \textcolor{blue}{\texttt{segment\_block}} \\

        \textbf{/* Reward Instruction */} \\
        Your task is to evaluate the relevance of each segment in answering the question below, to assist in identifying the segment(s) that most effectively answer the question. \\
        Question: \textcolor{blue}{\texttt{question}}\\
        Options: \textcolor{blue}{\texttt{options}}\\
        Treat the start and end frames of every sub-segment as cues for reconstructing what the segment might contain.
        Use these cues to judge how informative the segment is for answering the question and assign a score between 0\% and 100\%.
        Explain how the boundary frames shape your interpretation and why they lead you to the assigned relevance score.
        Please give the answer in the format: \texttt{\{"Segment \#": \{"explanation": str, "score": int\}\}} \\

        \rule[0.25\baselineskip]{\textwidth}{1pt} 
{\small
\textit{\textcolor{gray}{Placeholder Descriptions}} 
\begin{itemize}
    \setlength\itemsep{0.4em}
    \item[\textbullet] \textcolor{gray}{\texttt{frames\_number}: The number of sampled frames.}
    \item[\textbullet] \textcolor{gray}{\texttt{segment\_number}: The number of segments generated by the sampled frames, which equals the number of sampled frames plus one.}
    \item[\textbullet] \textcolor{gray}{\texttt{duration}: The total duration of the video in seconds. (e.g., ``180").}
    \item[\textbullet] \textcolor{gray}{\texttt{frame\_block}: A comma-separated string of sampled frame indices or timestamps (e.g., ``30, 60, 90, 120').}
    \item[\textbullet] \textcolor{gray}{\texttt{segment\_block}: A multi-line string defining the video segments. Each line follows  \\
    \texttt{Segment \#: [start, end]}.(e.g.,``Segment 0: [0s, 30s]\textbackslash\textbackslash Segment 1: [30s, 60s]\textbackslash\textbackslash ..."). }
    \item[\textbullet] \textcolor{gray}{\texttt{question}: The specific question to be answered about the video content.}
    \item[\textbullet] \textcolor{gray}{\texttt{options}: The options of the question to be answered about the video content.}
\end{itemize}
}
        \rule[0.25\baselineskip]{\textwidth}{1pt} 
        
        \textbf{Assistant} \\
        \textcolor{blue}{\texttt{\{"Segment 0": \{"explanation": "...", "score": ...\}, "Segment 1": \{...\}\}}}
        \end{minipage}
        \end{tabular}
    \end{tcolorbox}
    \vspace{-2mm}
\end{minipage}
\end{table*}

\begin{table*}[htbp]
\centering

\begin{minipage}{1\textwidth}
    \centering
    \caption{\textbf{Prompt for the reward model in subsequent rounds, with detailed placeholder descriptions.}}
\label{supp:tab_reward_following}
    \begin{tcolorbox}
        \centering
        \begin{tabular}{p{0.99\textwidth}}
        \begin{minipage}{0.99\textwidth}

        \textbf{User} \\
        \textbf{/* Task Description */} \\
        You are acting as a reward model in a multi-round video question-answering process. You have access to a \textcolor{blue}{\texttt{duration}}-frame video (\textcolor{blue}{\texttt{duration}} seconds), along with results from a previous round of evaluation. 
        In this round, one specific segment has been further divided to provide more detailed analysis. You are provided with \textcolor{blue}{\texttt{N}} new sampled frames to assess these sub-segments in relation to the question, at the following frame indices: \textcolor{blue}{\texttt{frame\_block}}. \\

        \textbf{/* Goal Question and Options*/} \\
        Question:\textcolor{blue}{\texttt{question}} \\
        Options:\textcolor{blue}{\texttt{options}} \\

        \textbf{/* Historical Segment Information */} \\
        In the last round, the video was divided into \textcolor{blue}{\texttt{candidate\_count}} segments, each evaluated for its relevance to the goal question.  
        Here are the results from all previous rounds: \\
        \textcolor{blue}{\texttt{historical\_block}} \\

        \textbf{/* Current Segment Information */} \\
        In this round, segment \textcolor{blue}{\texttt{parent\_label}} has been further explored with \textcolor{blue}{\texttt{frame\_number}} new uniformly sampled frames, dividing it into \textcolor{blue}{\texttt{segment\_number}} new sub-segments: \\
        \textcolor{blue}{\texttt{segment\_block}} \\

        \textbf{/* Reward Instruction */} \\
        Your task is to evaluate these new sub-segments for relevance to the original goal question based on provided frames, while considering the context and results from previous rounds. 
        Treat the start and end frames of every sub-segment as cues for reconstructing what the segment might contain.
        Use these cues to judge how informative the segment is for answering the question and assign a score between 0\% and 100\%.
        Explain how the boundary frames shape your interpretation and why they lead you to the assigned relevance score.
        Please respond in the format: \texttt{\{"Segment \#": \{"explanation": str, "score": int\}\}} \\

        \rule[0.25\baselineskip]{\textwidth}{1pt}

{\small
\textit{\textcolor{gray}{Placeholder Descriptions}} 
\begin{itemize}
    \setlength\itemsep{0.4em}
    \item[\textbullet] \textcolor{gray}{\texttt{frame\_number}: The number of sampled frames.}
    \item[\textbullet] \textcolor{gray}{\texttt{segment\_number}: The number of segments generated by the sampled frames, which equals the number of sampled frames plus one..}
    \item[\textbullet] \textcolor{gray}{\texttt{duration}: Duration of the video in seconds (e.g., ``180'').}
    \item[\textbullet] \textcolor{gray}{\texttt{frame\_block}: Comma-separated sampled frame timestamps or indices (e.g., ``30, 45, 60, 75'').}
    \item[\textbullet] \textcolor{gray}{\texttt{segment\_block}: A multi-line string containing new sub-segments formatted as {Segment \#: [start, end]}.}
    \item[\textbullet] \textcolor{gray}{\texttt{historical\_block}: A multi-line string containing prior segments, scores, and explanations from previous rounds.}
    \item[\textbullet] \textcolor{gray}{\texttt{parent\_label}: The label of the parent segment, e.g., ``1''.}
    \item[\textbullet] \textcolor{gray}{\texttt{candidate\_count}: Number of segments evaluated in the previous round.}
    \item[\textbullet] \textcolor{gray}{\texttt{question}: The question to be answered.}
    \item[\textbullet] \textcolor{gray}{\texttt{options}: The options of the question to be answered about the video content.}
\end{itemize}
}

        \rule[0.25\baselineskip]{\textwidth}{1pt}

        \textbf{Assistant} \\
        \textcolor{blue}{\texttt{\{"Segment 0": \{"explanation": "...", "score": ...\}, "Segment 1": \{...\}, ...\}}}

        \end{minipage}
        \end{tabular}
    \end{tcolorbox}
    \vspace{-2mm}
\end{minipage}
\end{table*}

\begin{table*}[htbp]
\centering

\begin{minipage}{1\textwidth}\vspace{3mm}
    \centering
    \caption{\textbf{Prompt for the agent in selection step, with detailed placeholder descriptions.}}
    \label{supp:tab_policy}
    \begin{tcolorbox}
        \centering

        \begin{tabular}{p{0.99\textwidth}}

        \begin{minipage}{0.99\textwidth}

        \textbf{User} \\
        \textbf{/* Task Description */} \\
        You are a helpful assistant with access to a video that is \textcolor{blue}{\texttt{duration}} frames long (\textcolor{blue}{\texttt{duration}} seconds).  \\
        You are tasked with exploring the video to gather the information needed to answer a specific question with complete confidence.  \\
        Question:\textcolor{blue}{\texttt{question}} \\
        Options:\textcolor{blue}{\texttt{question}} \\
        At each step, you may select one segment of the video to examine. Once you choose a segment, you will receive a set of representative frames sampled from that segment.  
        Use each exploration step strategically to uncover key details, progressively refining your understanding of the video’s content.  
        Continue exploring as needed until you have acquired all information necessary to answer the question.  \\
        In this round, you are provided with \textcolor{blue}{\texttt{memory\_count}} sampled frames stored in the memory module, with frame indices: \textcolor{blue}{\texttt{memory\_indices}}.  
        In the history exploration process, the video has been divided into \textcolor{blue}{\texttt{candidate\_total}} distinct segments, each covering a specific interval.  
        The interval and relevance score for each segment are detailed below. \\

        \textbf{/* Segment Information */} \\
        \textcolor{blue}{\texttt{candidate\_block}} \\

        \textbf{/* Exploration Instruction */} \\
        For each segment, we provide a fused score that adaptively combines two components to support your exploration:
        (1) an intrinsic reward, computed by an auxiliary video assistant based on the segment’s relevance to the question, and
        (2) a query score that reflects how many relevant clips are contained in the segment .
        Focus on the segments most likely to contain key information for confidently answering the question.  
        Now, proceed with your exploration, selecting the segment you wish to explore.  
        Please provide your choice in the following format: \texttt{\{Segment: int\}}. \\

        Before drawing a conclusion, examine the relevant details as thoroughly as possible to gather sufficient information. Every action you take should aim to deepen your understanding of the video, especially the parts related to the question. You have ample time, so focus on providing the most accurate answer possible.\\

        If you have enough information to answer the question, select the best answer from the options and directly provide the answer without giving any explanation. \\

        \rule[0.25\baselineskip]{\textwidth}{1pt}

{\small
\textit{\textcolor{gray}{Placeholder Descriptions}} 
\begin{itemize}
    \setlength\itemsep{0.4em}
    \item[\textbullet] \textcolor{gray}{\texttt{duration}: Duration of the video in seconds (e.g., ``180.0'').}
    \item[\textbullet] \textcolor{gray}{\texttt{question}: The question that the agent must eventually answer.}
    \item[\textbullet] \textcolor{gray}{\texttt{options}: The options of the question .}
    \item[\textbullet] \textcolor{gray}{\texttt{memory\_indices}: A comma-separated list of all frames stored in memory (e.g., ``30, 45, 60'').}
    \item[\textbullet] \textcolor{gray}{\texttt{memory\_count}: Number of stored frames.}
    \item[\textbullet] \textcolor{gray}{\texttt{candidate\_total}: Number of segments created in previous exploration rounds.}
    \item[\textbullet] \textcolor{gray}{\texttt{candidate\_block}: Multi-line description of each segment in the format \\ 
    \texttt{Segment \#: span=[start,end], score=float, explanation=str}.}
\end{itemize}
}

        \rule[0.25\baselineskip]{\textwidth}{1pt}

        \textbf{Assistant} \\
        \textcolor{blue}{\texttt{\{"Segment": int\}}}

        \end{minipage}
        \end{tabular}
    \end{tcolorbox}
    \vspace{-2mm}
\end{minipage}
\end{table*}

\begin{table*}[htbp]

\centering
\begin{minipage}{1\textwidth}\vspace{3mm}
    \centering
    \caption{\textbf{Prompt for discovering semantic queries from user instruction.}}
    \label{supp:tab_query_generation}
    \begin{tcolorbox}
        \centering

        \begin{tabular}{p{0.99\textwidth}}

        \begin{minipage}{0.99\textwidth}

        \textbf{User} \\
        \textbf{/* Role */} \\
        Produce short text queries for a VideoCLIP-style retriever. \\

        \textbf{/* Input */} \\
        ONE multiple-choice question about a video (with options). \\
        Question: \textcolor{blue}{\texttt{question}} \\
        Options: \textcolor{blue}{\texttt{options}} \\

        \textbf{/* Goal */} \\
        Do \emph{not} answer the question. Convert the question into 1--5 stand-alone semantic queries  
        that can be fed directly into the text encoder to retrieve relevant clips. \\

        \textbf{/* Output Format */} \\
        - Return only a JSON array of strings, length 1--5, no extra text. \\
        - Each query must contain 6--12 lowercase words, concise and concrete. \\

        \textbf{/* Writing Rules */} \\
        1) Prefer copying key phrases from the question/options; avoid adding specific names, places,  
        colors, or timestamps that are not present in the input. \\
        2) If the question includes a temporal anchor (e.g., ``after the interview with xxx''),  
        include that anchor verbatim. \\
        3) Each query should be a compact description: [temporal anchor if any] + [target from options] + [simple action or neutral cue]. \\
        4) No duplicates. If fewer high-quality queries are possible, output fewer. \\

        \textbf{/* Example Format Only (not content) */} \\
        Reply strictly in JSON format as: \\
        \texttt{\{"query1": "...", "query2": "...", ...\}} \\
        with no additional text. \\

        \rule[0.25\baselineskip]{\textwidth}{1pt}

{\small
\textit{\textcolor{gray}{Placeholder Descriptions}}
\begin{itemize}
    \setlength\itemsep{0.4em}
    \item[\textbullet] \textcolor{gray}{\texttt{question}: The multiple-choice question to be rewritten into simple retriever queries.}
    \item[\textbullet] \textcolor{gray}{\texttt{options}: The answer options from which key phrases may be reused.}
\end{itemize}
}

        \rule[0.25\baselineskip]{\textwidth}{1pt}

        \textbf{Assistant} \\
        \textcolor{blue}{\texttt{[\ "query1", "query2", ...\ ]}}

        \end{minipage}
        \end{tabular}
    \end{tcolorbox}
    \vspace{-2mm}
\end{minipage}
\end{table*}

\begin{table*}[htbp]
\centering

\begin{minipage}{1\textwidth}\vspace{3mm}
    \centering
    \caption{\textbf{Prompt for extracting new semantic queries from current frames.}}
    \label{supp:tab_query_update}
    \begin{tcolorbox}
        \centering
        \hspace{-6mm}
        \begin{tabular}{p{0.99\textwidth}}
        \hspace{1mm}
        \begin{minipage}{0.99\textwidth}

        \textbf{User} \\
        You are a video-understanding assistant. \\

        \textbf{/* Input Information */} \\
        - Frames with timestamps: \textcolor{blue}{\texttt{time\_of\_frames}} \\
        - A multiple-choice question with options \\
        Question: \textcolor{blue}{\texttt{question}} \\
        Options: \textcolor{blue}{\texttt{options}} \\
        - Historical semantic queries information (already known): \\
        \textcolor{blue}{\texttt{history\_queries}} \\

        \textbf{/* Task */} \\
        From the \textbf{current frames only}, extract \emph{new}, concrete semantic queries that can guide subsequent retrieval or exploration toward answering the question. \\

        \textbf{/* Strict Rules */} \\
        1) Output only short, concrete semantic queries (nouns or verb-noun phrases with less than 10 words). No full sentences. \\
        2) Each query must be directly grounded in the provided frames and must \emph{not} appear in the historical information. \\
        3) Avoid generic words (``scene'', ``shot'', ``clip'') and avoid speculation (no unseen colors, names, or places). \\
        4) Provide 2--5 items. If no new cues exist, return an empty dict \texttt{\{\}}. \\
        5) Prefer salient, discriminative tokens that are easy to search (objects, OCR snippets, logos,  
           tools, distinctive props, on-screen text, gestures, sound-indicated events). \\

        \textbf{/* Output Format */} \\
        Reply strictly in JSON as: \\
        \texttt{\{"query1": "...", "query2": "...", ...\}} \\
        with \textbf{no} extra text. \\

        \vspace{1mm}
        \textbf{/* Negative Example (do NOT do this) */} \\
        - [``Frame 6 shows the villain in a shattered mirror environment with broken glass pieces around''] \\

        \rule[0.25\baselineskip]{\textwidth}{1pt}

{\small
\textit{\textcolor{gray}{Placeholder Descriptions}}
\begin{itemize}
    \setlength\itemsep{0.4em}
    \item[\textbullet] \textcolor{gray}{\texttt{time\_of\_frames}: A list of frame timestamps  (e.g., ``0s, 1s, 2s, 3s''.}
    \item[\textbullet] \textcolor{gray}{\texttt{history\_info}: Historical semantic queries extracted from previous rounds. The new output must \textbf{not} contain any repetition.}
\end{itemize}
}

        \rule[0.25\baselineskip]{\textwidth}{1pt}

        \textbf{Assistant} \\
        \textcolor{blue}{\texttt{\{"query1": "...", "query2": "...", ...\}}}

        \end{minipage}
        \end{tabular}
    \end{tcolorbox}
    \vspace{-2mm}
\end{minipage}
\end{table*}

\end{document}